\documentclass[conference]{IEEEtran}
\IEEEoverridecommandlockouts
\usepackage{cite}
\usepackage{amsmath,amssymb,amsfonts}
\usepackage{graphicx}
\usepackage{textcomp}
\usepackage{xcolor}
\def\BibTeX{{\rm B\kern-.05em{\sc i\kern-.025em b}\kern-.08em
    T\kern-.1667em\lower.7ex\hbox{E}\kern-.125emX}}
    
\usepackage{fancyhdr}
\usepackage{xcolor,soul,framed}
\usepackage{xspace}
\usepackage{array}
\usepackage{eqparbox}
\usepackage{url}
\usepackage{longtable}
\usepackage{lipsum}
\usepackage{blindtext}
\usepackage{makecell}
\usepackage{mathtools}
\usepackage{commath}
\usepackage{multirow}
\usepackage{tabularx}
\usepackage[normalem]{ulem}
\usepackage{xspace}
\usepackage{algorithm}
\usepackage{algpseudocode}
\usepackage{amsfonts}
\usepackage{placeins}
\usepackage{amssymb}
\usepackage{multirow}
\usepackage{tabularx}
\usepackage{caption}
\usepackage{subcaption}
\usepackage{pifont}

\usepackage{xspace}
\usepackage{algorithm}
\usepackage{algpseudocode}
\usepackage{amsfonts}
\usepackage{placeins}
\usepackage{booktabs}
\usepackage[normalem]{ulem}
\definecolor{green}{RGB}{0, 150, 00}
\definecolor{orange}{RGB}{255, 120, 0}
\def\BibTeX{{\rm B\kern-.05em{\sc i\kern-.025em b}\kern-.08em
    T\kern-.1667em\lower.7ex\hbox{E}\kern-.125emX}}

\usepackage[pagebackref=true,breaklinks=true,colorlinks,bookmarks=false]{hyperref}
\hypersetup{
    colorlinks=true,
    filecolor=magenta,      
    urlcolor=magenta,
}

\usepackage{fancyhdr}
\begin{document}

\title{Analysis of Semi-Supervised Methods for Facial Expression Recognition}
\author{\IEEEauthorblockN{Shuvendu Roy, Ali Etemad}
\IEEEauthorblockA{Dept. ECE and Ingenuity Labs Research Institute \\
Queen's University, Kingston, Canada\\
\{shuvendu.roy, ali.etemad\}@queensu.ca}}

\maketitle
\thispagestyle{fancy} 

\begin{abstract}
Training deep neural networks for image recognition often requires large-scale human annotated data. To reduce the reliance of deep neural solutions on labeled data, state-of-the-art semi-supervised methods have been proposed in the literature. Nonetheless, the use of such semi-supervised methods has been quite rare in the field of facial expression recognition (FER). In this paper, we present a comprehensive study on recently proposed state-of-the-art semi-supervised learning methods in the context of FER. We conduct comparative study on \textit{eight} semi-supervised learning methods, namely Pi-Model, Pseudo-label, Mean-Teacher, VAT, MixMatch, ReMixMatch, UDA, and FixMatch, on three FER datasets (FER13, RAF-DB, and AffectNet), when various amounts of labeled samples are used. We also compare the performance of these methods against fully-supervised training. Our study shows that when training existing semi-supervised methods on as little as 250 labeled samples per class can yield comparable performances to that of fully-supervised methods trained on the full labeled datasets. To facilitate further research in this area, we make our code publicly available at: \href{run:https://github.com/ShuvenduRoy/SSL\_FER}{https://github.com/ShuvenduRoy/SSL\_FER}.

\end{abstract}


\section{Introduction}\label{sec:intro}

Facial expressions play an important role in human communications. As a result, growing efforts are being made toward developing facial expression recognition (FER) methods that can facilitate better human-machine interaction systems. Real-world applications of FER systems include driving assistants \cite{leng2007experimental}, personal mood management systems \cite{thrasher2011mood, sanchez2013inferring}, health-care assistants \cite{tokuno2011usage}, emotion-aware multimedia \cite{cho2019instant}, and others. Although FER systems have shown great promise and improvements over the past few years, the FER remains challenging due to several factors such as occlusions, illuminations, scene backgrounds, challenging viewing angles, ethnicity, and demographic factors. Recently, deep learning solutions have shown the potential to effectively perform FER and solve many of these problems \cite{sepas2021capsfield,CL_MEx,kolahdouzi2021face,sepas2020facial}. However, deep learning models require very large human-annotated datasets to achieve their optimal performance, and collecting such large datasets with human annotations is a costly and time-demanding process. Hence, FER solutions capable of learning robust representations from a relatively small amount of labeled data are highly desired.

\begin{figure}[t]
\centering
\includegraphics[width=0.49\textwidth]{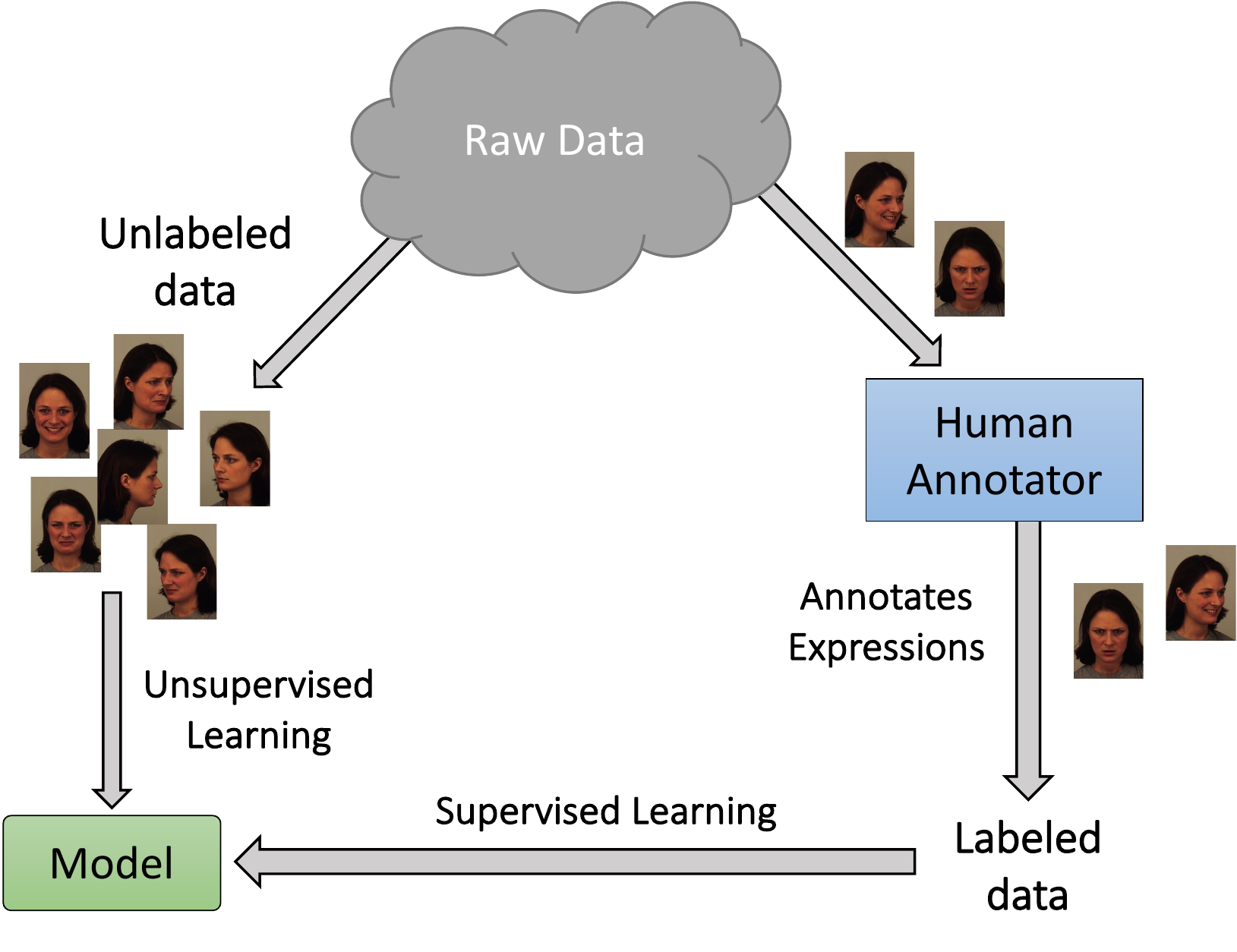} 
\caption{Overview of semi-supervised learning for FER.}
\label{fig:embedding}
\end{figure}

To deal with the unavailability of large annotated datasets, methods such as self-supervised and semi-supervised learning leverage unlabeled data to learn important features with minimal supervision. While self-supervised methods \cite{ST_CLR, pourmirzaei2021using, CL_MEx} learn from pseudo-labels generated from unlabeled data, semi-supervised learning methods \cite{jiang2021boosting,wang2021multi} take a more hybrid approach to utilize small amounts of labeled data to guide the learning of large unlabeled datasets. Despite the success of these methods in image and video representation learning, their use for FER has been less explored. A high-level overview of semi-supervised learning for FER is shown in Fig \ref{fig:embedding}.

\begin{figure*}[t]
\centering
\includegraphics[width=1\textwidth]{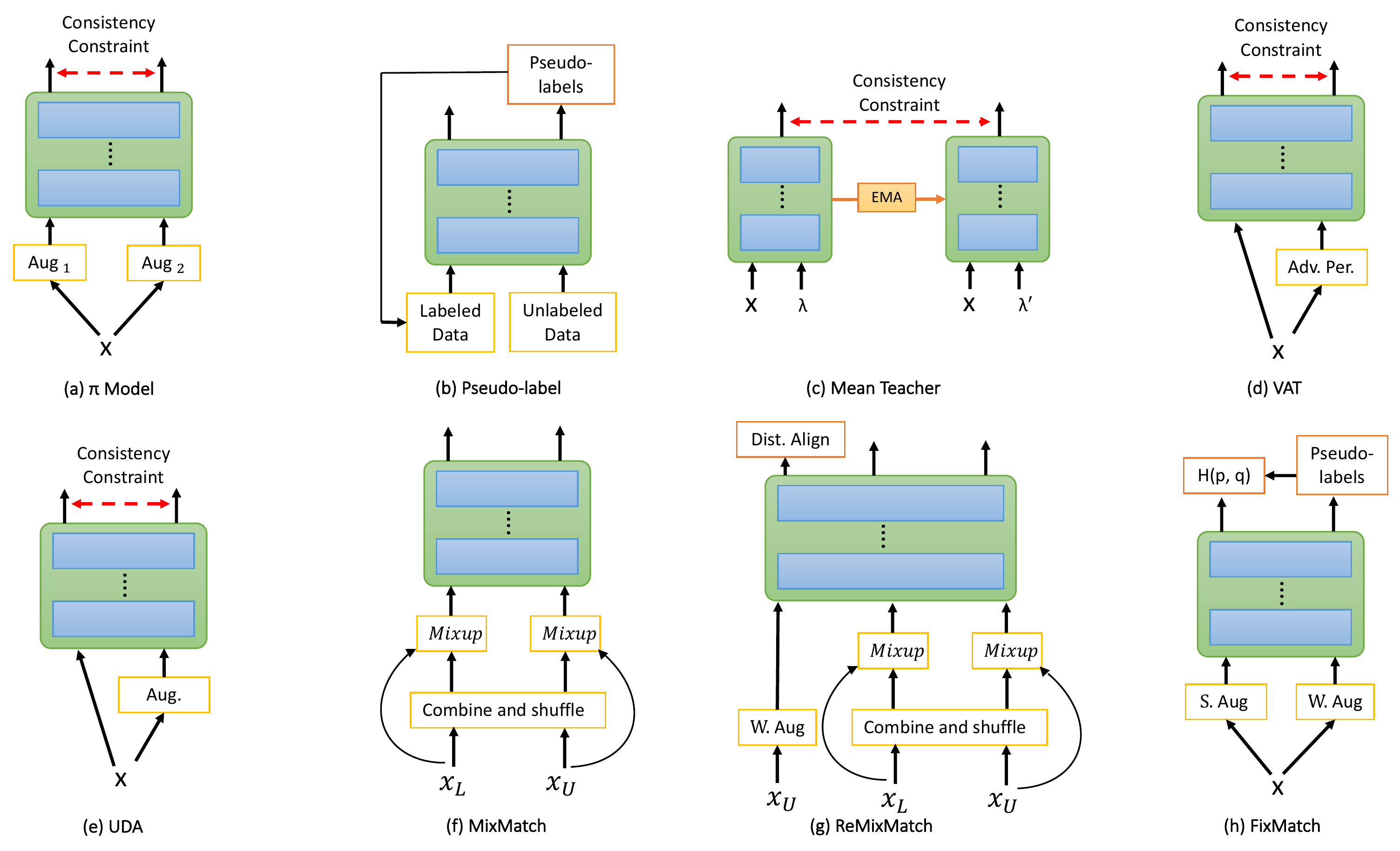}
\caption{Overview of the eight state-of-the-art semi-supervised methods studied in this paper for FER. Here, Aug., W. Aug, and S. Aug represent augmentations, weak, and augmentations respectively. Adv. Per. represents adversarial perturbation. EMA is the exponential moving average of the original model.}
\label{fig:models}
\end{figure*}

Recently, a wide range of semi-supervised learning methods have been proposed which can be categorized into two broad categories: pseudo-labeling \cite{pseudo_labels} and consistency regularization \cite{pi_model, uda, vat}. In pseudo-labeling \cite{pseudo_labels}, a model trained on the labeled data is used to predict the labels for the unlabelled data. The predicted pseudo-labels with higher confidence are subsequently used as labels for the unlabeled data. The model is then trained with the unlabeled data in a supervised setting using the generated pseudo-labels. Consistency regularization-based methods are based on the assumption that a realistic perturbation (e.g. augmentation) on an image does not change its semantics, and therefore the output of the model should remain unchanged. The most common type of these methods reduces the distance between the embeddings of two augmentations of an unlabeled image \cite{pi_model, uda}. More recently, consistency regularization has been combined with pseudo-labeling in a hybrid framework that shows impressive performance in a wide variety of tasks \cite{fixmatch,remixmatch,zhang2021flexmatch}. An example of such a hybrid method is FixMatch \cite{fixmatch}, which first applies a weak augmentation for an unlabeled image and predicts the pseudo-label for it. Then the model is trained with a hard augmentation of the same image using the predicted pseudo-label. 

Although there has been some progress in semi-supervised learning in FER, to the best of our knowledge, there is no comprehensive and comparative study on the recent semi-supervised methods applied to FER. In this work, we study eight recently proposed semi-supervised methods on three popular facial expression datasets. The aim of this work is to study the adaptability and performance of such methods on FER without any specialization in expression recognition. We conduct extensive studies with different amounts of labeled data and find impressive results with as little as only 10 labeled samples per class used for training. 
These methods, when trained with as few as 250 samples per class, show competitive results to the fully-supervised method trained on the same dataset when all of the sample labels are used. 

In this paper we make the following contributions:
\begin{itemize}
    \item We present a comprehensive study on eight recent semi-supervised learning methods on three popular FER datasets.
    
    \item We conduct a comparative analysis of the methods and further compare them with fully supervised learning using the same backbone encoder. 
    
    \item To facilitate quick reproduction and further research on the topic of semi-supervised FER, we release the code for this work which contains the implementations of all the methods in this study. 
\end{itemize}

The remainder of this paper is organized as follows. In the next section, we describe the 8 semi-supervised methods in detail, namely Pi-Model, Pseuro-label, Mean-Teacher, VAT, MixMatch, ReMaxMatch, UDA, and FixMatch. Following, we describe the experiments including the description of the datasets, a comparison of the performances of the semi-supervised methods, and the impact of parameters. Finally, we present the concluding remarks.

\section{Semi-supervised Methods}\label{sec:method}
\begin{figure*}[ht]
\centering
\includegraphics[width=0.90\textwidth]{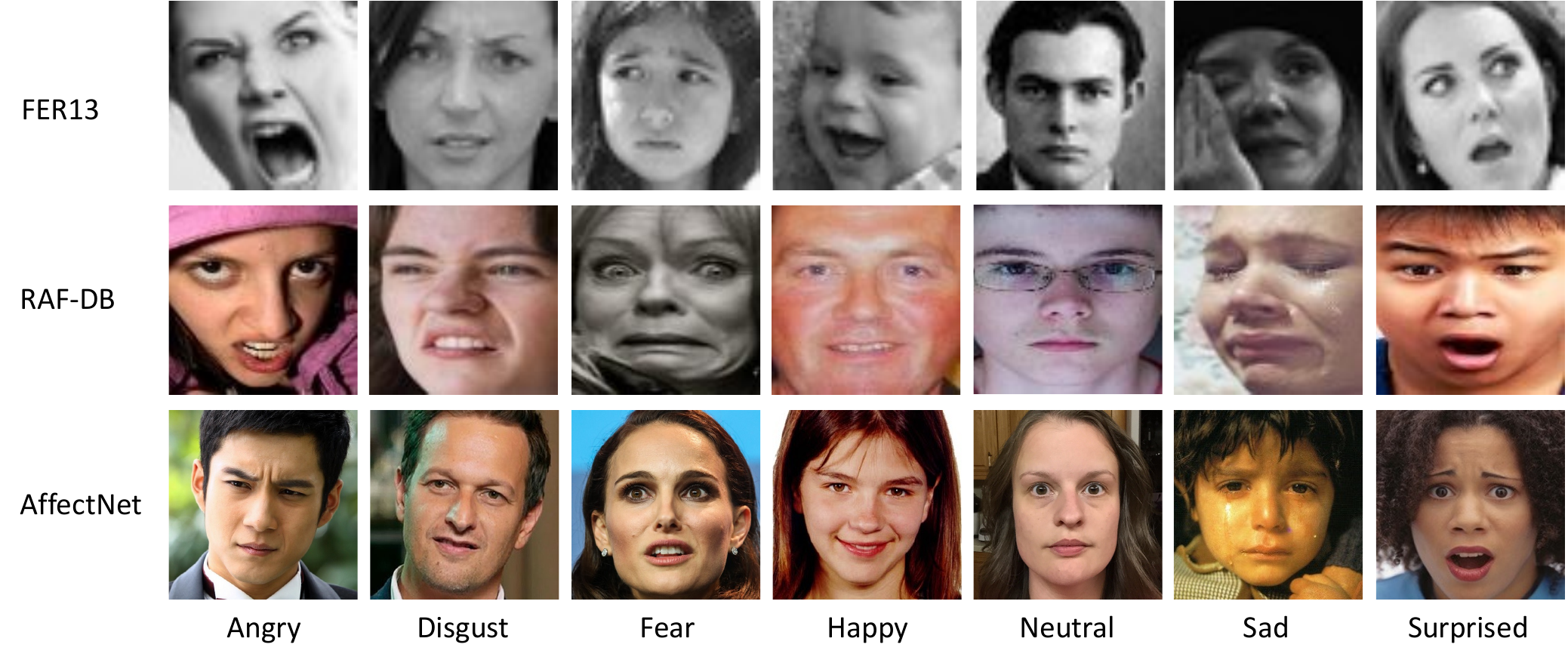}
\caption{Examples of images from FER13, RAF-DB, and AffectNet datasets.}
\label{fig:example}
\end{figure*}

In this section, we describe the problem setup, and an overview of 8 popular and effective semi-supervised methods for image representation learning.

\subsection{Problem Setup}

Assume we have a small labeled dataset $D_l=\{(x_i^l,y_i^l)\}^N_{i=1}$ containing $N$ images and its corresponding class labels. Let's assume we also have a large unlabeled dataset $D_u=\{(x_i^u)\}^M_{i=1}$, where $M\gg N$. Although we do not have any annotated labels for the images in the unlabeled dataset $D_u$, our aim is to utilize $D_u$ and $D_l$ together to help the model learn better representations. In this problem setup, we assume a separate validation dataset $D_v=\{(x_i^u,y_i^u)\}^V_{i=1}$ that is used to test the final performance of the model. We ensure no overlap between the images in labeled, unlabeled, and validation sets, i.e., $D_l \cap D_u \cap D_v = \emptyset $.

\subsection{Semi-Supervised Methods}

We study eight recent semi-supervised methods namely: Pi-Model, Pseuro-label, Mean-Teacher, VAT, MixMatch, ReMaxMatch, UDA, and FixMatch. Following, we present an overview of each of these methods.

\subsubsection{Pi-Model}
Pi-Model \cite{pi_model} is one of the most popular \textit{Consistency Regularization} based semi-supervised learning methods. In this approach, the model generates two augmentations of an image from the unlabelled data, and a regularization loss function reduces the difference in their learned embeddings. At the same time, the pi-model is trained on the labeled data with regular cross-entropy loss in a supervised way. The model also introduces stochastic behavior in the prediction of the model using dropout, random max-pooling, and a randomized augmentation module. The structure of the Pi-Model is shown in Figure \ref{fig:models} (a). The loss on the unsupervised data is represented as
\begin{equation}
		\mathbb{E}_{x\in D_l} \mathcal{R}(f(\theta, \tau_1(x)),f(\theta, \tau_2(x))),
\end{equation}
where $\tau_1$ and $\tau_2$ are two random augmentations applied on the input image $x$, $f$ is the encoder network, $\theta$ represents the parameters of the model $f$, and $\mathcal{R}$ is the consistency regularization function.

\subsubsection{Mean Teacher}
Mean teacher \cite{mean_teacher} is built on the concept of consistency regularization, similar to the approach proposed for the Pi-model. However, instead of using the same encoder to generate embeddings of two augmented images, the mean teacher uses an exponential moving average (EMA) of the encoder to predict the embedding for the second image. The regular encoder is called a \textit{student model}, whereas the EMA of the student model is called a \textit{teacher model}. The mean teacher applies the regularization loss on the prediction of the teacher and student models on two augmentations of the same image. A visual illustration of the mean teacher method is depicted in Figure \ref{fig:models} (c). The loss function proposed for the Mean teacher can be represented as
\begin{equation}
		\mathbb{E}_{x\in D_u} \mathcal{R}(f(\theta, \tau_1(x)),f(EMA(\theta), \tau_2(x))),
\end{equation}
where $EMA(\theta)$ is the teacher model. All other notations are similar to that of the Pi-model. The formula for updating the EMA teacher model from the student model is represented as 
\begin{equation}
    \theta'= m \theta'+ (1-m)\theta,
\end{equation}
where $m$ is smoothing coefficient for the EMA update.

\subsubsection{VAT}
Virtual Adversarial Training (VAT) \cite{vat} is conceptually similar to the Pi-model. While the Pi-model regularizes the embedding of two augmentations on the same image, VAT introduces the concept of adversarial attack as an alternate to augmentation. More specifically, it generates an adversarial transformation of the input. Following, consistency regularization is applied to the input image and the transformed image. A visual illustration of VAT is shown in Figure \ref{fig:models} (d). The loss function for VAT is represented as follows
\begin{equation}
		\mathbb{E}_{x\in D_u} \mathcal{R}(f(\theta, x),f(\theta, \gamma^{adv}(x))),
\end{equation}
where $\gamma^{adv}$ adversarial perturbation operator.

\subsubsection{UDA}
Unsupervised domain adaptation (UDA) \cite{uda} is another consistency regularization method that showed a large improvement in the performance by replacing the regular augmentation module with recently proposed hard augmentation techniques such as AutoAugment \cite{autoaugment} and RandAugment \cite{randaugment} that generate very dynamic and diverse augmentations of an input image.
Figure \ref{fig:models} (e) presents the diagram of the UDA method that is very similar to Pi-model and VAT, except for the augmentation module which is replaced with hard augmentations. The loss function of the UDA can be represented as
\begin{equation}
		\mathbb{E}_{x\in D_u} \mathcal{R}(f(\theta, x),f(\theta, \tau(x))),
\end{equation}
where $\tau$ is the hard augmentation module.

\subsubsection{Pseudo-label}
Pseudo-label \cite{pseudo_labels} proposed a very simple yet efficient solution for semi-supervised learning. The concept of this method acts as the fundamental block for many of the current state-of-the-art methods.
In Pseudo-label, the model predicts the output class probability for each of the unlabeled data which is considered as a pseudo-label for the unlabeled image. The model is then trained in a supervised setting using the labels for the labeled data and the pseudo-labels for the unlabeled data. A visual illustration of the Pseudo-label method is shown in Figure \ref{fig:models} (b). The loss function for the Pseudo-label method can be represented as
\begin{equation}
		\mathcal{L}=\mathcal{L}(y_i^l, f(\theta, x_i^l))+\lambda  \mathcal{L}(y_i^{u},f(\theta, x_i^u)),
	\end{equation}
where $y_i^{u}$ is the prediction pseudo-label for the unlabeled image $x_i^{u}$, and $\lambda$ is a coefficient that balances the impact of the two loss functions.

\subsubsection{MixMatch}
MixMatch \cite{mixmatch} is a hybrid semi-supervised learning method that unifies consistency regularization with pseudo-labeling. Like the pseudo-labeling methods, MixMatch also predicts the pseudo-labels for the unlabeled data and utilizes them to train the model in supervised settings along with the labeled data. However, the new distinctive component of this algorithm is a Mixup operation that generates mixed inputs and mixed labels by interpolating between the labeled and unlabeled images and their corresponding labels. The Mixup operation of MixMatch can be represented with the following formula:
\begin{equation}
    x'=\alpha x_l + (1-\alpha)x_u,
\end{equation}
where $x_l$ and $x_u$ are the labeled and unlabeled input images, and $\alpha$ balances influence of unlabeled images on the generated mixed image. The value of $\alpha$ is randomly sampled from a beta distribution. 

MixMatch also introduces the concept of generating multiple instances of an unlabeled image with multiple augmentations. More specifically, MixMatch applies one augmentation on the labeled data $(x_i^{l},y_i^{l})$, but $k$ weak augmentations on an unlabeled image $x_j^u$ to generate $k$ instances,
whose predictions are then averaged to generate one pseudo-label $\hat{y_j}$ for each unlabeled image.

The Mixup operation on a batch of labeled data ($d_l \in D_l$) and unlabeled data ($d_u \in D_u$) generates the dataset $d_l^\prime$ and $d_u^\prime$. Accordingly, the combined MixMatch loss function of the labeled and unlabeled data can be represented as:
\begin{align}
    \mathcal{L}_l &= \frac{1}{|d_l^\prime|} \sum_{x, y \in d_l^\prime} H(y, f(x, \theta)) \label{eqn:l_x} , \\
    \mathcal{L}_u &= \frac{1}{C|d_u^\prime|} \sum_{x^\prime, y^\prime \in d_u^\prime} \|y\prime - f(x, \theta)\|_2^2 \label{eqn:l_u} , \\
    \mathcal{L} &= \mathcal{L}_l + \lambda \mathcal{L}_u . \label{eqn:l_combined}
\end{align}
Here, $H(p,q)$ is the cross-entropy between the distribution $p$ and $q$, $C$ is the number of classes, and $\lambda$ is a hyper-parameter to balance the influence of the labeled and unlabeled loss terms. A visual illustration of the MixMatch method is shown in Figure \ref{fig:models} (f).

\subsubsection{ReMixMatch}
ReMixMatch \cite{remixmatch} is an extension of MixMatch with two new ideas: distribution alignment and augmentation anchoring. The distribution alignment encourages the distribution of the predictions on the unlabeled data to be similar to that of the labeled data. Augmentation anchoring is added as a replacement for the consistency regularization of MixMatch to encourage the representation of strongly augmented images to be similar to that of weakly augmented images. Here the augmentation anchoring technique uses one weakly augmented image against multiple strongly augmented images. The method also introduced a new strong augmentation method called CTAugment that is more suitable in semi-supervised learning settings. The ReMixMatch method is illustrated in Figure \ref{fig:models} (g).

\subsubsection{FixMatch}
FixMatch \cite{fixmatch} is a unified framework that combines consistency regularization with pseudo-labeling in a very simplified way to generate a very simple semi-supervised learning framework. For an unlabeled image, FixMatch first applies weak augmentations and predicts the pseudo-labels for them. Next, it applies a hard augmentation on the same images and trains the model in supervised settings with pseudo-labels. FixMatch only uses the pseudo-labels if the confidence of the predictions is higher than a pre-defined threshold ($p_{\text{cutoff}}$). FixMatch uses only standard shift and flip augmentations for its weak augmentation module. For the hard augmentation module, it uses RandAugment \cite{randaugment}, CTAugment \cite{remixmatch}, and CutOut \cite{cutout}. A visual illustration of FixMatch is depicted in Figure \ref{fig:models} (h).

\begin{table*}[ht]
    \centering
    \setlength
    \tabcolsep{3.6pt}
    \caption{The performance of different semi-supervised methods on FER13, RAF-DB, and AffectNet datasets, when only 10, 25, 100, and 250 labeled samples are used for training. Here, bold and underline represent the best and second best accuracy for each setting. }
    \begin{tabular}{l|rrrr|rrrr|rrrr}
    \toprule
        & \multicolumn{4}{c|}{\textbf{FER13}} & \multicolumn{4}{c|}{\textbf{RAF-DB}}  & \multicolumn{4}{c}{\textbf{AffectNet}}  \\
        \cmidrule(l{3pt}r{3pt}){1-13}
        Method / $m$                & 10 labels        & 25 labels      & 100 labels     & 250 labels    & 10 labels         & 25 labels      & 100 labels     & 250 labels    & 10 labels         & 25 labels      & 100 labels     & 250 labels      \\
       \cmidrule(l{3pt}r{3pt}){1-13}

        $\Pi$-model	&37.09  &40.87& 50.66 & 56.42  &39.86	 &50.97& 63.98 & 71.15&24.17&25.37&31.24&32.40 \\	
        
        Mean Teacher	& 45.21	& \underline{55.14}& 52.17 & 58.06 &62.05  & 45.17& 45.57 & \textbf{76.85}	& 19.54 & 20.21 & 20.80 & 44.05\\	
        
        VAT	& 24.95	 &\textbf{55.22} &51.55& 55.64 & \underline{63.10}  & 45.82& 62.05 & 59.45 & 17.68 & \underline{35.02} & 37.68 & 37.92 	\\	
        
        UDA	& \underline{46.72}	& 49.89 & 50.62  & \underline{60.68} & 46.87&\textbf{53.15}  & 58.86 & 60.82 & 27.42& 32.16 & 37.25 & 37.64  \\	
        
        Pseudo-label	&32.79 &36.04 & 49.21 & 54.88 & 58.31     & 39.11   &54.07  & 67.40 & 18.00&21.05&33.05&37.37	 	\\
        
        MixMatch	& 45.69	& 46.41 & \underline{ 55.73} &  58.27 & 36.34& 43.12 &	\underline{64.14} & 73.66 &  \textbf{30.80}&32.40&39.77&\underline{48.31} \\	
        
        ReMixMatch	&41.07& 43.25 & 44.62  &57.49& 37.35& 42.56	 & 42.86 & 61.70 & 29.28&33.54& \underline{41.60} &46.51 	 	\\	
        
        FixMatch	&\textbf{47.88} & 49.90 & \textbf{59.46} & \textbf{62.20} & \textbf{63.25} & \underline{52.44} & \textbf{64.34} & \underline{75.51}&  \underline{30.08}  &  \textbf{38.31} & \textbf{46.37} & \textbf{51.25}	 	\\	      

        \bottomrule
    \end{tabular}
    \label{tab:results}
\end{table*}

\begin{table}[]
\caption{Fully supervised training with full and partial training data, as well as the top semi-supervised performance.}
\begin{center}
\begin{tabular}{l|cc | c}
\toprule
& \multicolumn{2}{c|}{\textbf{Fully-sup.}} & \textbf{Semi-sup. (best)} \\
\cmidrule(l{3pt}r{3pt}){1-4}
\textbf{Dataset} & \textbf{All labeled data}  & \textbf{250 labels/class} & \textbf{250 labels/class} \\ \cmidrule(l{3pt}r{3pt}){1-4}
FER13 & 64.57 & \textcolor{orange}{53.58} &  \textcolor{green}{62.20}\\
RAF-DB & 80.47 & \textcolor{orange}{65.87} & \textcolor{green}{76.85} \\
AffectNet & 54.91 & \textcolor{orange}{40.28} & \textcolor{green}{51.25}\\

\bottomrule
\end{tabular}
\label{tab_sup}
\end{center}
\end{table}

\section{Experiments and Performance}\label{sec:results}

In this section, we first describe the three datasets used to evaluate the eight state-of-the-art semi-supervised methods. We then present the implementation setup in detail, including the encoder architecture and training protocol. We then present the results. Finally, we present a sensitivity analysis of the semi-supervised methods with respect to their hyper-parameters.

\subsection{Datasets}

\subsubsection{FER13 \cite{fer13}}
This is a facial expression dataset that has been collected and labeled automatically by the Google image search API. All the images in this dataset are re-scaled to a resolution of $48\times48$. The dataset contains over 35K images of 7 basic expressions where 28K images belong to the training split. Figure \ref{fig:example} (top) shows a few examples of the dataset.

\subsubsection{RAF-DB \cite{raf_db}}
This is an in-the-wild dataset that contains around 15K images with 12K images on the training split. A total of 315 human annotators worked on the annotation of this dataset and each label is assured to be annotated by around 40 individual annotators. This dataset also contains 7 expression classes. Some examples of RAF-DB dataset is shown in Figure \ref{fig:example} (middle).

\subsubsection{AffectNet \cite{affectnet}}
This is a very large-scale in-the-wild dataset collected from the Internet with three search engines and expression keywords. We took the images for 7 basic expressions (a total of around 284K images), similar to the FER13 and RAF-DB datasets. Some examples of the AffectNet dataset are shown in Figure \ref{fig:example} (bottom).

\subsection{Implementation details}
Here we present the implementation details including the encoder and the training protocol used in this study. Unless mentioned otherwise, we use the same setup for all the results reported in this paper.

\subsubsection{Encoder}
We use ResNet \cite{resnet} as the general backbone encoder for all the semi-supervised methods, as our preliminary experiments showed that this style architecture outperforms others such as VGG-style networks \cite{vgg}. More specifically, we use the ResNet-18 as our default encoder which is trained with an input resolution of $224\times224$. A variant of ResNet called WideResNet, which has been specially designed to work well with images of very low resolution, is used in our study for the FER13 dataset.

\subsubsection{Training protocol}
For semi-supervised training, we first randomly select $N=n\times C$ images for the labeled dataset $D_l$, where $C$ is the total number of classes and $n$ is the number of labeled images to be taken par class. In this study, we present the result for all the methods with $n \in \{10, 25, 100, 250 \}$ samples. Following FixMatch, we have trained all the models for $2^{20}$ iterations with a batch-size of 64 for the labeled data and a 7-times larger batch size for the unlabeled data. The cutoff confidence value for selecting a pseudo-label is set to 0.95 for all the pseudo-label based methods. The models are trained with SGD optimizer with a momentum of 0.9 and an initial learning rate of 0.03. A cosine learning rate decay is used as a scheduler to reduce the learning rate over the training iterations. Weight decay is used as a regularizer with a value of 0.0005. For the EMA model, a moving average weight of 0.999 is used. For the models with sharpening distribution, a temperature value of 0.5 is utilized.

\begin{figure*}
    \input{ACII/figures/aug_example}
\end{figure*}

\begin{figure*}
   \centering
     
     \begin{subfigure}[b]{0.32\textwidth}
         \centering
         \includegraphics[width=1.25\textwidth]{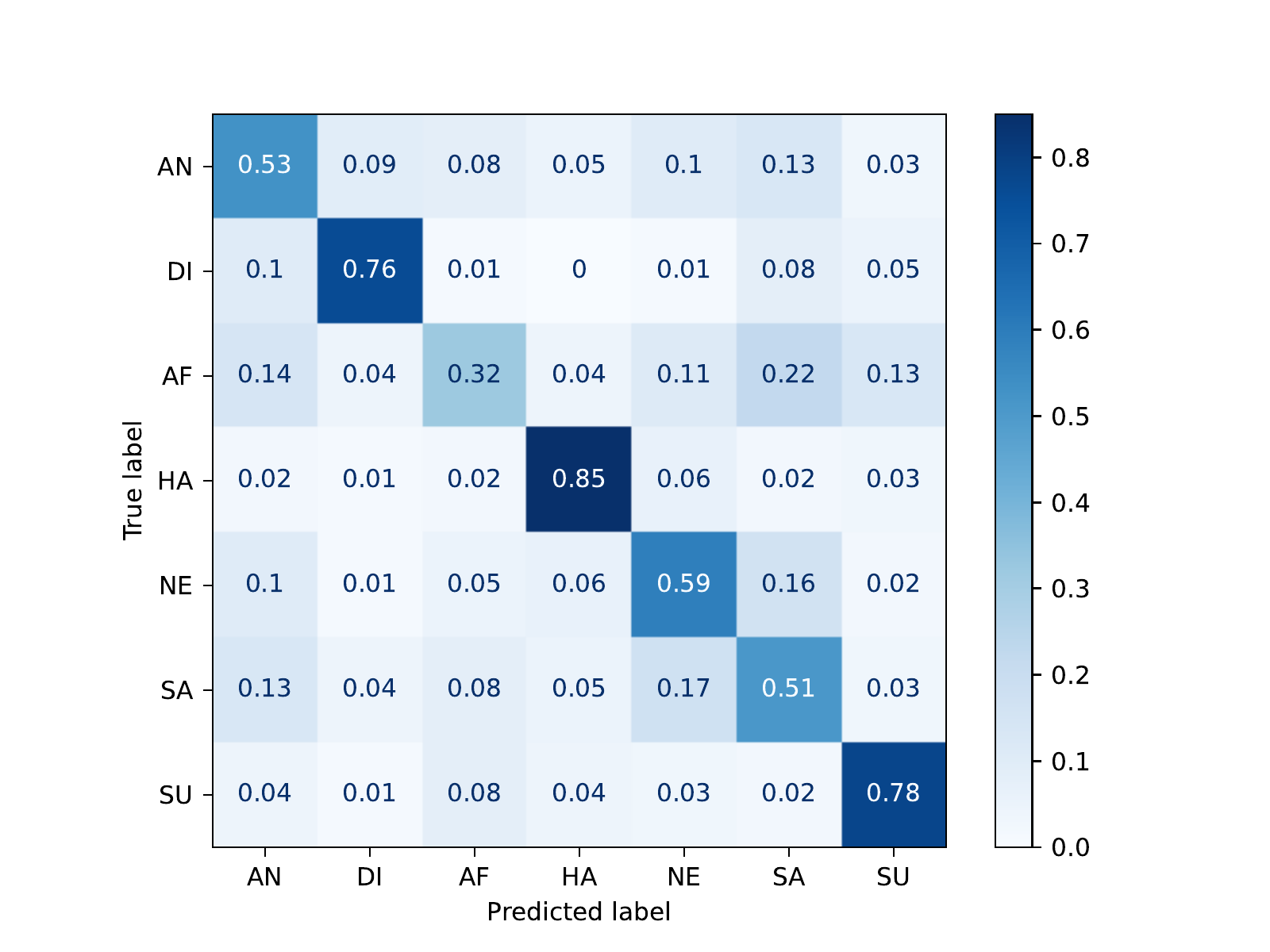}
         \caption{\textit{FER13 (FixMatch)}}
         \label{fig:con_fer}
     \end{subfigure}
     \hfill
     \begin{subfigure}[b]{0.32\textwidth}
         \centering
         \includegraphics[width=1.25\textwidth]{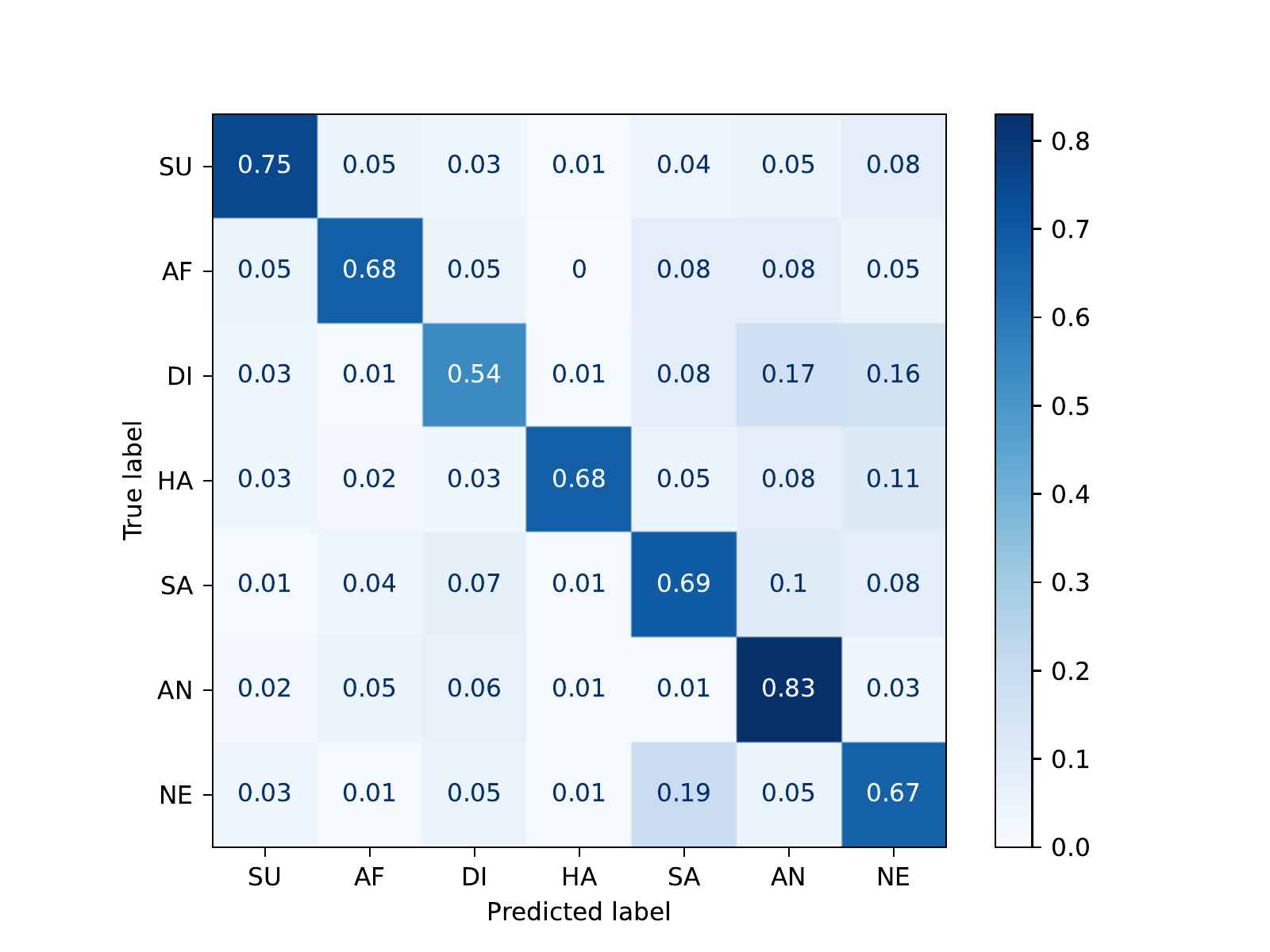}
         \caption{\textit{RAF-DB (Mean Teacher)}}
         \label{fig:con_raf_db}
     \end{subfigure}
      \hfill
     \begin{subfigure}[b]{0.32\textwidth}
         \centering
         \includegraphics[width=1.25\textwidth]{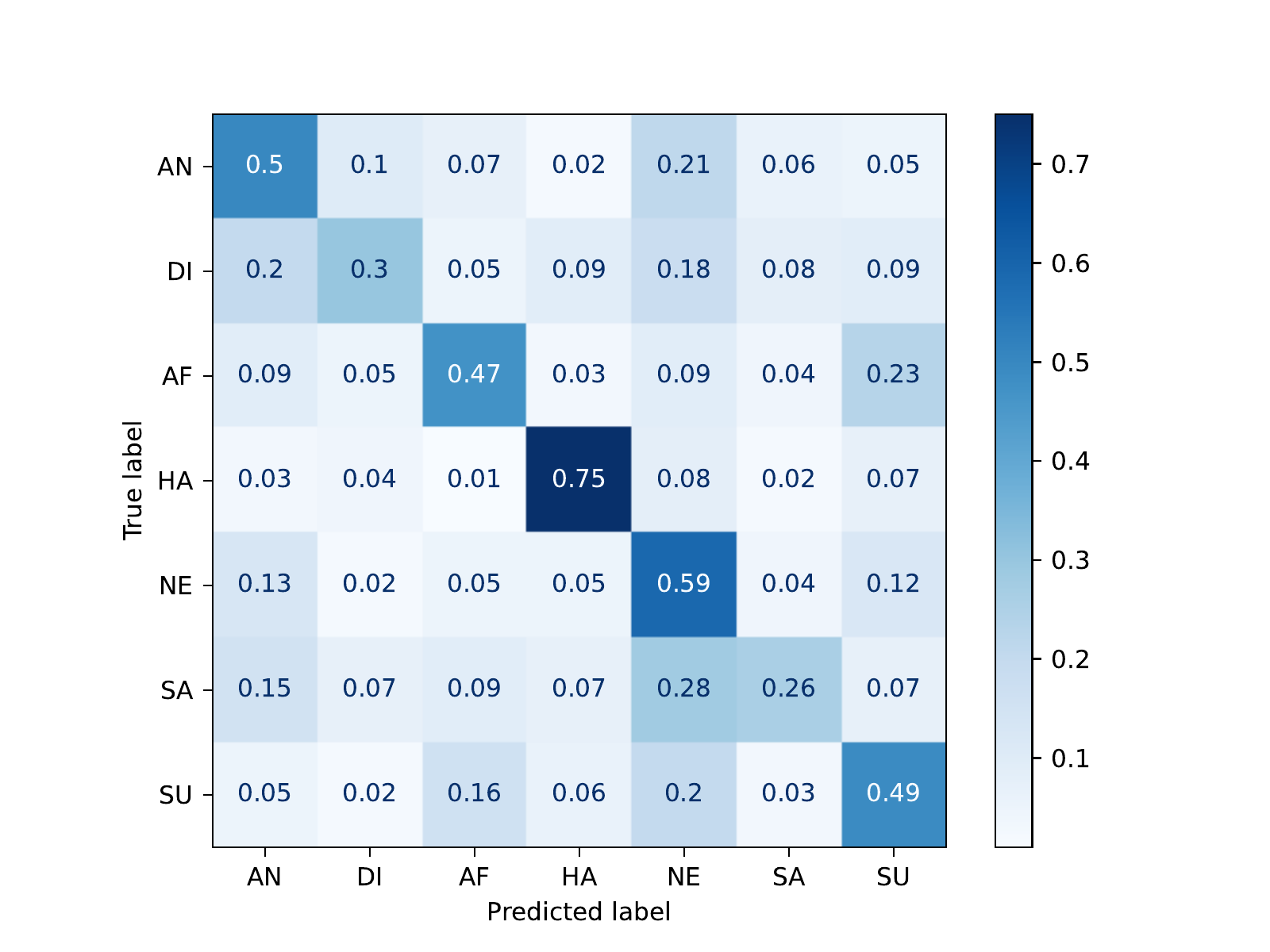}
         \caption{\textit{AffectNet (FixMatch)}}
         \label{fig:con_affect}
     \end{subfigure}
    
    \caption{Confusion matrices for the best semi-supervised model on FER13, RAF-DB, and AffectNet datasets.}
    \label{fig:consution}
\end{figure*}

To compare the performance of the semi-supervised methods, we also train the same backbone encoder in a fully-supervised setting. For the experiment with fully supervised training, we present the results in two settings: (1) using the full training data for training, and (2) using 250 images per class for training (maximum amount of labeled data used by the semi-supervised methods). For a fair comparison, we keep all the training settings the same where applicable. For fully supervised training, the augmentation module consists of random resizing, random horizontal flip, and random crop. The fully supervised models are also trained for $2^{20}$ iterations with SGD optimizer and a cosine learning rate decay scheduler. All the methods are trained with the proposed weak and strong augmentations in the original method. Some example of weak and strong augmentations applied to a facial image is illustrated in Figure \ref{fig:aug}.

\subsection{Results}
Table \ref{tab:results} presents the results for the eight semi-supervised methods on FER13, RAF-DB, and AffectNet datasets. Note that, the experiments for Pseudo-label, Mean Teacher, MixMatch, and FixMatch are done with the best hyper-parameters for each dataset that are found from the sensitivity study which will be presented in Section \ref{sec:sensitivity}. As standard practice in semi-supervised literature \cite{mixmatch,fixmatch}, we present the results for 10, 25, 100, and 250 labels per class, i.e., a total of 70, 175, 700, and 1750 labeled images for training. For the FER13 dataset, these training settings represent only 0.2\%, 0.6\%, 2.0\%, and 6.0\% of the total training data respectively, while for RAF-DB these training numbers represent 0.5\%, 1.4\%, 5.7\% and 14.2\% of the total training data. Finally for AffectNet, we end up using 0.03\%, 0.07\%, 0.28\% and 0.7\% of the total training set. Our experiments in this table show that for FER13, FixMatch outperforms the other methods (shown in bold) when 10, 100, and 250 labeled samples are used, whereas VAT performs the best when 25 labels are included in the training. In terms of the second best (shown with underline), there is no clear pattern among the different methods. As expected, the more labeled samples are used during training, the better the performance becomes (250 labeled samples $>$ 100 labeled samples $>$ 25 labeled samples $>$ 10 labeled samples).
For RAF-DB, FixMatch performs strongly again by showing the best results for 10 and 100 labeled samples, while being the second-best approach for 25 and 250 labeled samples. UDA and Mean Teacher exhibit the best results when 25 and 250 labeled samples are used, respectively. Finally, AffectNet shows a similar trend where FixMatch outperforms the other solutions when 25, 100, and 250 labeled samples are used. When only 10 labels are used, FixMatch proves to be the second best method following MixMatch as the best method.

Overall, the result from Table \ref{tab:results} shows that out of 12 experiments (3 datasets, 4 label settings), FixMatch outperforms the others in 8 instances, while achieving the second best in 3 other settings. MixMatch obtains the next best results outperforming the others in 1 instance and being the second best in 3 more instances. 3 other top results are achieved by VAT, UDA, and Mean Teacher each with one best result among the 12 settings. 

We summarize the results above in Table \ref{tab_sup} by presenting the best results (best semi-supervised method trained with 250 labeled samples) from Table \ref{tab:results}. We also present the results of fully-supervised training with all the labeled data available during training, as well as fully-supervised training when only 250 labeled samples are present at the training. Here we observe that for FER13,
the semi-supervised solution is only 2.4\% less than the model trained in a fully-supervised fashion with all the training data but still 8.6\% better than training with the same amount of labels but with a fully-supervised setup (not the semi-supervised approach). 
For RAF-DB the semi-supervised method is only 3.6\% less than the model trained on the full dataset, but a considerable 11.0\% better than the supervised model trained on the same amount of labeled data.
Finally, for AffectNet, the semi-supervised approach is 3.7\% lower than the fully-supervised model while using 140 times less labeled data. This performance is 11\% better than the fully-supervised model trained with the same amount of labeled samples.

\begin{figure*}
    \centering
     \begin{subfigure}[b]{0.24\textwidth}
         \centering
         \includegraphics[width=1.1\textwidth]{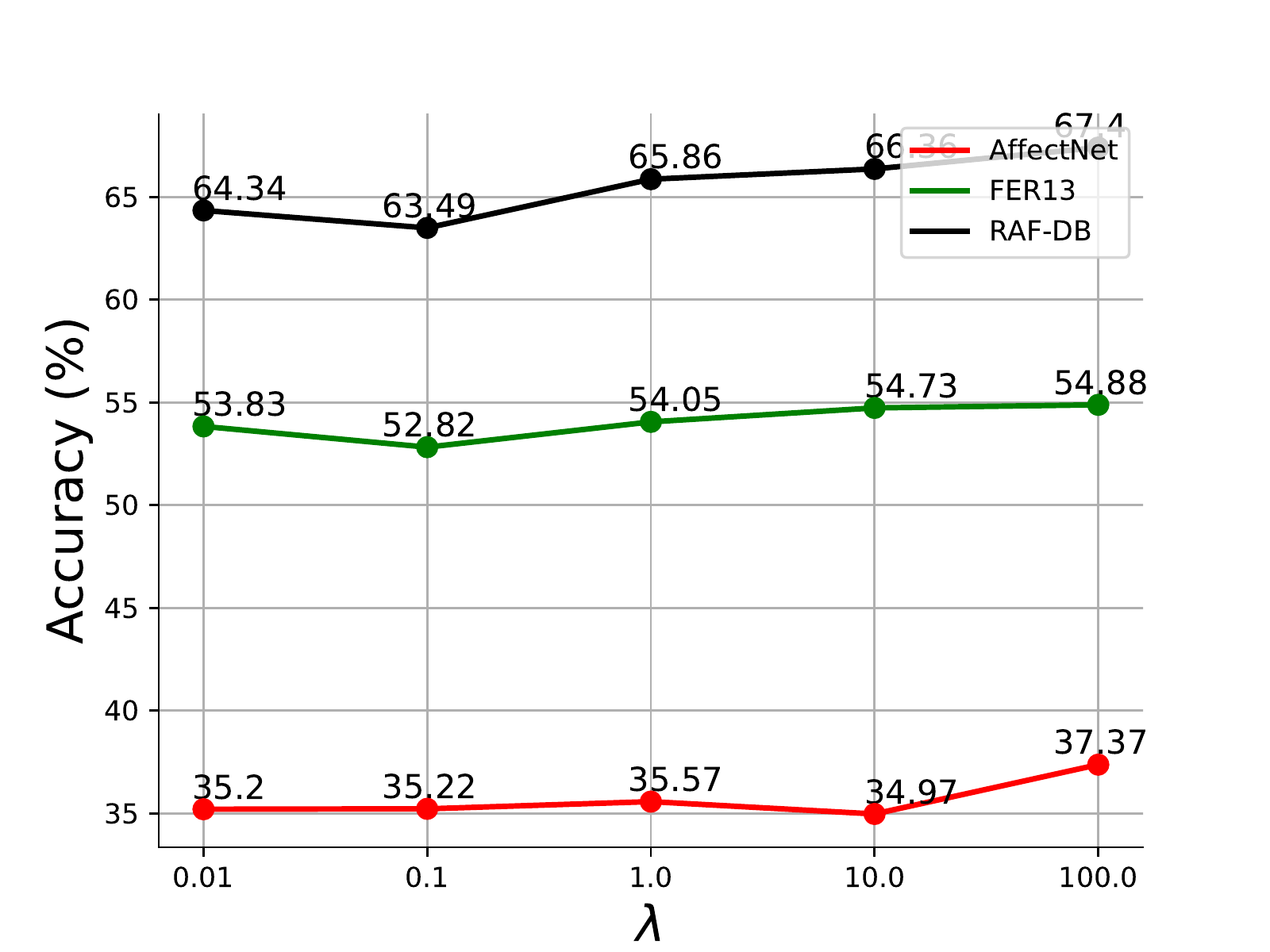}
         \caption{Accuracy vs. $\lambda$ for the \textit{\textbf{Pseudo-label}} method}
         \label{fig:lambda_pseudolabel}
     \end{subfigure}
     \hfill
      \begin{subfigure}[b]{0.24\textwidth}
         \centering
         \includegraphics[width=1.1\textwidth]{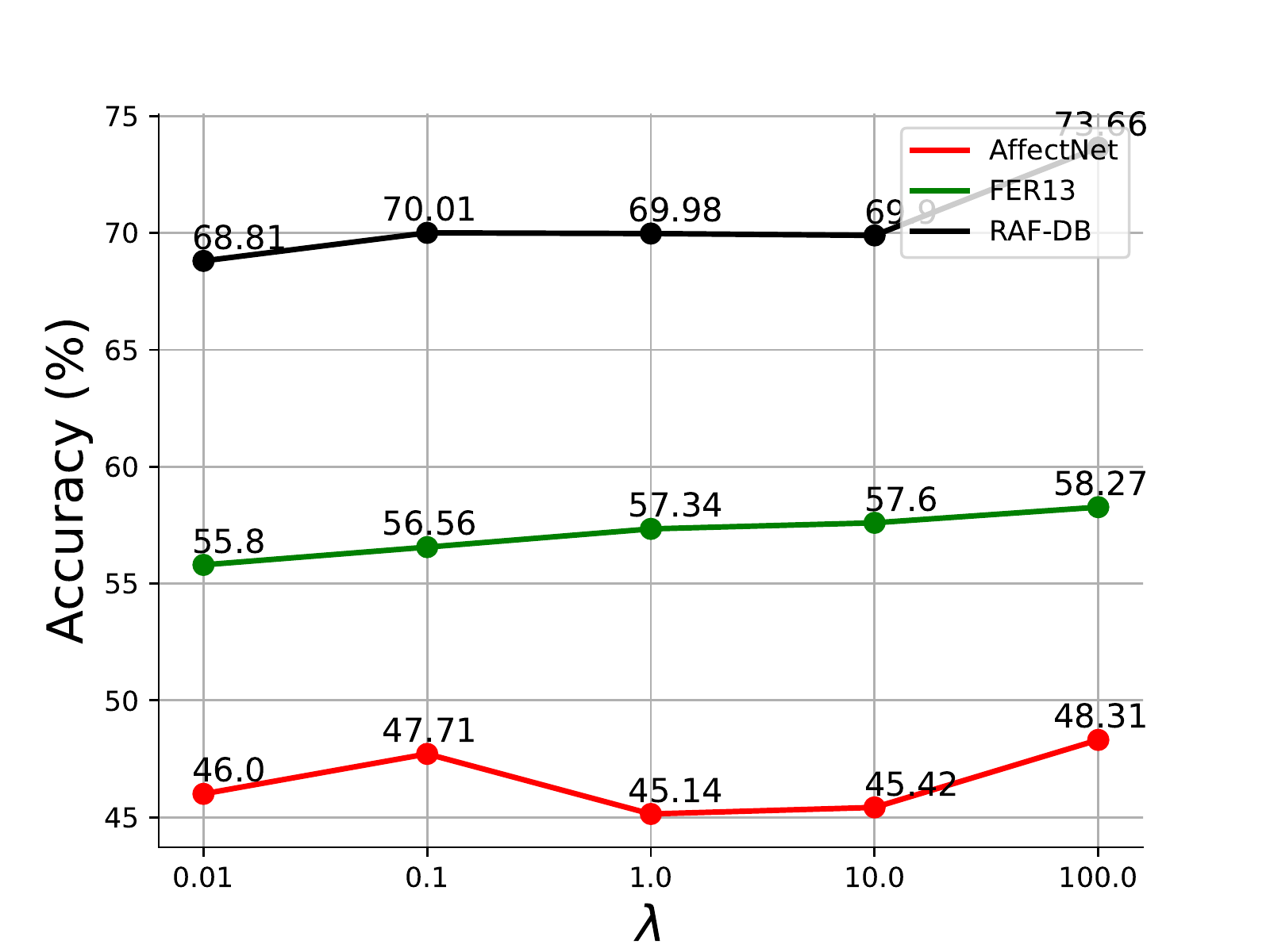}
         \caption{Accuracy vs. $\lambda$ for the \textit{\textbf{MixMatch}} method.}
         \label{fig:lambda_mixmatch}
     \end{subfigure}
    \hfill
     \begin{subfigure}[b]{0.24\textwidth}
         \centering
         \includegraphics[width=1.1\textwidth]{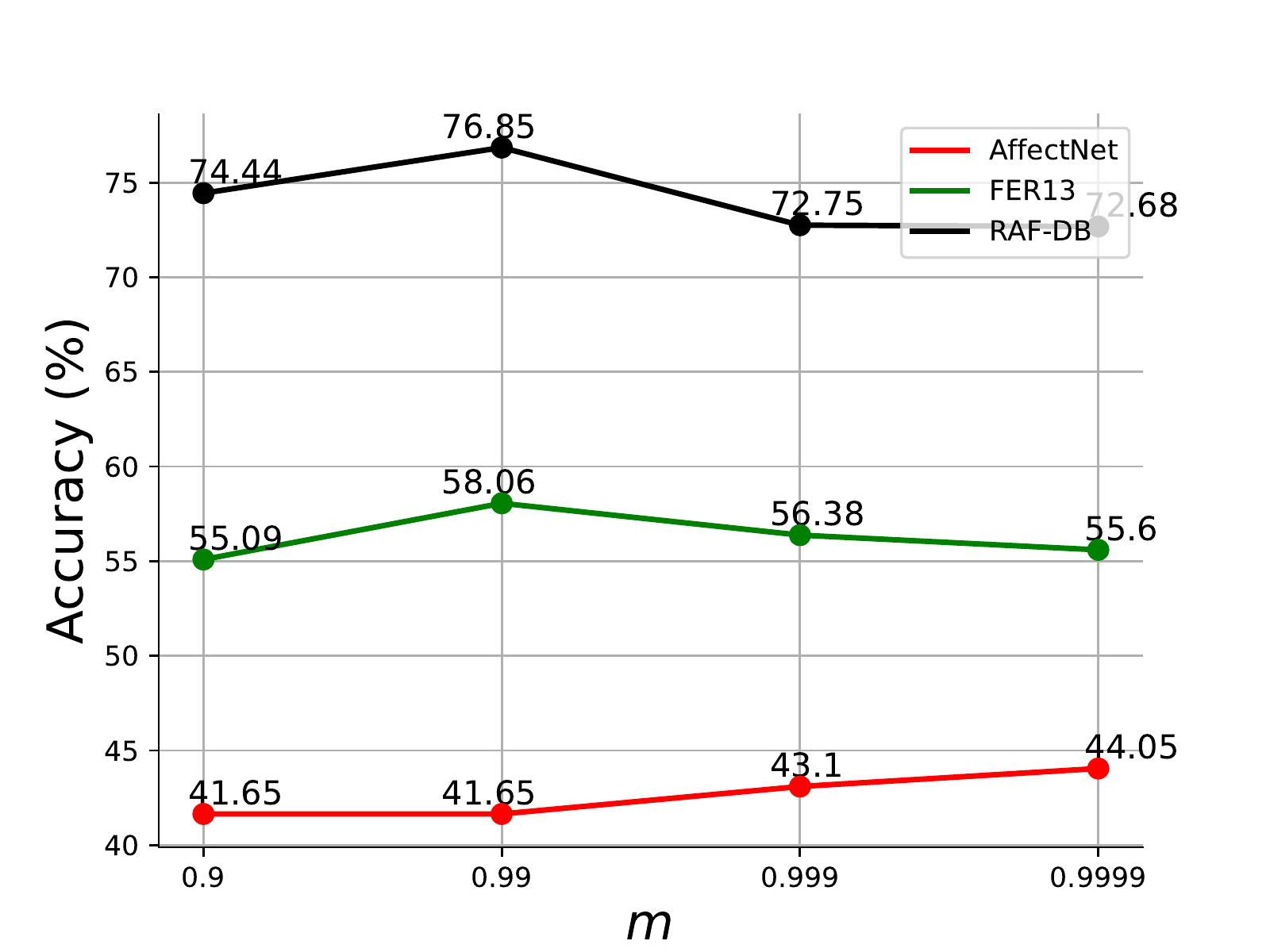}
         \caption{Accuracy vs. $m$ of EMA for the \textit{\textbf{Mean Teacher}} method.}
         \label{fig:m_meanteacher}
     \end{subfigure}
      \hfill
     \begin{subfigure}[b]{0.24\textwidth}
         \centering
         \includegraphics[width=1.1\textwidth]{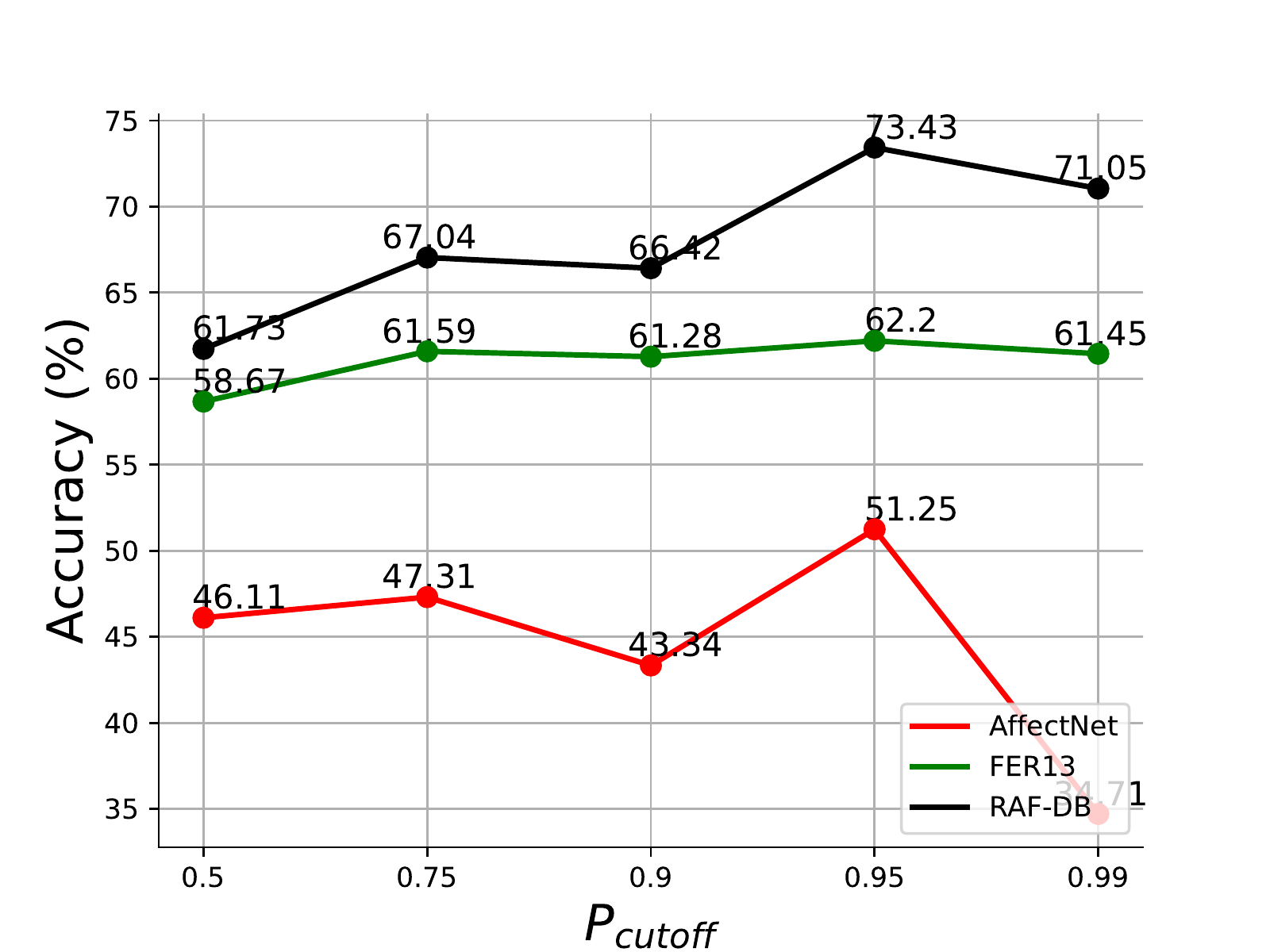}
         \caption{Accuracy vs. $p_{\text{cutoff}}$ for the \textit{\textbf{FixMatch}} method.}
         \label{fig:cutoff_fixmatch}
     \end{subfigure}
    
    \caption{Sensitivity study of various parameters for different semi-supervised methods on all three datasets.}
    \label{fig:sensitivity}
\end{figure*}

To further analyze the performance of the models, we evaluate the confusion matrices of the predictions of the best semi-supervised models on test data for each dataset respectively, and show the results in Figure \ref{fig:consution}. Our analysis shows that even with training the models with small amounts of labeled data (250 samples), the semi-supervised approaches can learn to make strong predictions with reasonable mistakes. Specifically, we observe that for FER13 (Figure \ref{fig:con_fer}), the majority of the mistakes occur by predicting `sad' instead of `afraid'. In RAF-DB (Figure \ref{fig:con_raf_db}), on other hand, `neutral' was often misclassified as `sad', while for AffectNet (Figure \ref{fig:con_affect}), `sad' conversely classified as `neutral' the most.

\subsection{Sensitivity study}\label{sec:sensitivity}
In this work, we adopt state-of-the-art semi-supervised methods for the task of FER where we use the training setups suggested in the original papers. However, since the problem domain of FER is quite different from that of general object classification (which is what the majority of the original semi-supervised literature has focused on), we also explore some of the important hyper-parameters of these methods. Specifically, we present a sensitivity study on the $\lambda$ value of Pseudo-label and MixMatch, $m$ for the EMA model of Mean Teacher, and the $p_{\text{cutoff}}$ value of FixMatch. The other semi-supervised solutions do not have any major hyper-parameters to tune. The experiments are done on all three datasets (FER13, RAF-DB, and AffectNet), and the results are presented in Figure \ref{fig:sensitivity}. In all cases, 250 labeled samples are used. 

As shown in Figure \ref{fig:lambda_pseudolabel}, the experiment on the $\lambda$ parameter of the Pseudo-label method shows improvement for higher values of $\lambda$, which means giving higher importance to the unsupervised loss term results in better accuracy. 
In the original paper \cite{pseudo_labels}, a $\lambda$ of $1.0$ was used as the default value for general object classification on the CIFAR10 dataset. However, our experiment shows that for FER, even higher values of $\lambda$ can be used which will further improve the results.
We observed a maximum improvement of around 3\% accuracy for AffectNet with this optimal value of $\lambda$ when compared to the default value. 
As shown in Figure \ref{fig:lambda_mixmatch}, a similar trend is also observed for $\lambda$ in MixMatch, where a maximum improvement of approximately 4\% is observed on RAF-DB with higher emphasis on the unlabeled loss term. 

The smoothing coefficient $m$ of the EMA in the Mean Teacher method is known to impact the performance of this semi-supervised solution. In the original paper \cite{mean_teacher}, higher values of $m$ (close to $1$) were suggested for optimal performance. In general, $m = 0.99$ works well on object classification, which as depicted in Figure \ref{fig:m_meanteacher}, also gives the best accuracy for FER13 and RAF-DB datasets. However, AffectNet shows better performance with even higher values of $m$. 

Finally, we present the sensitivity study on $p_{\text{cutoff}}$ value for FixMatch in Figure \ref{fig:cutoff_fixmatch}. The experiment shows that the best performance is achieved with a cutoff probability of $0.95$ across all the datasets, which is in line with the observations in the original FixMatch method \cite{fixmatch}. A large drop is observed for higher and lower cutoff values on all three datasets.

\section{Summary}
This paper presented a comprehensive study on different semi-supervised methods for FER using recent and state-of-the-art semi-supervised methods that have been originally proposed for general computer vision tasks. In particular, we adopted Pi-Model, Pseudo-label, Mean-Teacher, VAT, MixMatch, ReMixMatch, UDA, and FixMatch techniques in the context of FER and compared their performance against each other as well as fully-supervised settings on three popular datasets (FER13, RAF-DB, and AffectNet). Our study showed that even when the semi-supervised methods use a small portion of labeled data, they can achieve performances that are very competitive to the fully supervised methods trained using the full labeled dataset. We also compared the results with supervised models that use the same amount of labeled training data to that of semi-supervised methods, and observed significant improvements for the semi-supervised methods. Finally, we studied the impact of different hyper-parameters used in the semi-supervised methods to obtain a better understanding of the optimum settings for semi-supervision in FER. This work will serve as a baseline for future research on FER using semi-supervised approaches.

\section*{Acknowledgements}
We would like to thank BMO Bank of Montreal and Mitacs for funding this research. We are also thankful to SciNet HPC Consortium for helping with the computation resources.

\bibliographystyle{IEEEtran}
\bibliography{IEEEabrv,ref}
\end{document}